%% file: iclr2026_conference.tex
\documentclass{article} % For LaTeX2e
\usepackage{iclr2026_conference,times}

\input{math_commands.tex}

\usepackage{hyperref}
\usepackage{url}
\usepackage{fontawesome5}
\definecolor{darkpink}{RGB}{255, 20, 147}
\usepackage{enumitem}
\usepackage{amssymb}
\definecolor{ETHBlue}{RGB}{33,92,175}   %
\definecolor{ETHGreen}{RGB}{98,115,19}      %
\definecolor{ETHPurpleDark}{RGB}{140,10,89} %
\definecolor{ETHPurple}{RGB}{163,7,116} %
\definecolor{ETHGray}{RGB}{111,111,111} %
\definecolor{ETHPurpleDark}{RGB}{183,53,45}    %
\definecolor{ETHPetrol}{RGB}{0,120,148} %
\definecolor{ETHBronze}{RGB}{142,103,19}    %
\definecolor{ETHOrange}{RGB}{230, 100, 50}
\usepackage{graphicx}
\usepackage{subcaption}
\usepackage{tikz}
\usepackage{xcolor}
\usetikzlibrary{arrows.meta,positioning,fit,backgrounds}

\title{Loop as a Bridge: Can Looped Transformers Truly Link Representation Space and Natural Language Outputs?}

\iclrfinalcopy

\author{
\textbf{Guanxu Chen}$^{1,2}$\quad
\textbf{Dongrui Liu}$^{1}$\quad
\textbf{Jing Shao}$^{1}\thanks{Corresponding Author.}$\quad
\vspace{1em}\\
$^1$ Shanghai Artificial Intelligence Laboratory, 
$^2$ Shanghai Jiao Tong University,\\
\it\footnotesize ~~lm.cgx@sjtu.edu.cn\quad\quad\{liudongrui, shaojing\}@pjlab.org.cn
}

\begin{document}

\maketitle

\begin{abstract}
Large Language Models (LLMs) often exhibit a gap between their internal "knowledge" and their explicit linguistic outputs. In this report, we empirically investigate whether \textbf{Looped Transformers (LTs)}—architectures that increase computational depth by iterating shared layers—can bridge this gap by utilizing their iterative nature as a form of introspection. Our experiments reveal that while increasing loop iterations narrows the gap, it is partly driven by a degradation of their internal "knowledge" carried by representations. Moreover, another empirical analysis suggests that current LTs' ability to perceive representations does not improve across loops; it is only present in the final loop. These results suggest that while LTs offer a promising direction for scaling computational depth, they have yet to achieve the introspection required to truely link representation space and natural language.
\end{abstract}

\section{Introduction}
Recent advancements in Large Language Models (LLMs) have shifted the focus from direct response to complex multi-step reasoning \citep{gemini,guo2025deepseek,o1_card}. Key techniques such as Chain-of-Thoughts (CoTs, \citet{nye2021show,wei2022chain}) allow LLMs to refine and verify their answers through intermediate reasoning steps \citep{shao2024deepseekmath,liu2025understanding}. This phenomenon implies a relationship between \textcolor{ETHPetrol}{}{reasoning accuracy} and \textcolor{ETHGreen}{}{verification capabilities}: LLMs' ability to produce a correct solution is often upper-bounded by their ability to verify the correctness of their own solution. 
In parallel, another line of LLM research—encompassing AI Monitoring \citep{templeton2024scaling,liu2024efficient,liu2025latent} and Introspection \citep{lindsey2025emergent}—has moved beyond analyzing mere textual outputs to designing internal monitors based on model representations and interpretability. Even when LLMs are linguistically ambiguous or inconsistent to verify their own response \citep{madsen2024self,turpin2024language,chen2025reasoning}, their internal representations often exhibit high separability. This indicates that \textcolor{ETHPurpleDark}{}{the information carried by LLMs' representations}—their implicit "knowledge" and "awareness"—often surpasses \textcolor{ETHGreen}{}{their explicit linguistic output for verification}.

Synthesizing above observations, we identify a hierarchical phenomenon where \textcolor{ETHPurpleDark}{}{the knowledge and information encapsulated in LLMs' representations} cannot be fully expressed during \textcolor{ETHGreen}{}{their linguistic verification}, while \textcolor{ETHPetrol}{}{LLMs' action and decision} depend heavily on such \textcolor{ETHGreen}{}{self-verification}. Over the past year, progress in reasoning has gradually bridged the reliability gap between self-verification and LLMs' action through scaling sequence length and computation \citep{shao2024deepseekmath,liu2025understanding}. However, the gap between their internal representation and verification output remains.

To address this problem, this report turns the attention to \textbf{Looped Transformers (LTs)}. Unlike traditional single-pass, feed-forward architectures, LTs adopt a recursive way, reusing weights to iterate internal representations multiple times within the network. To bridge the gap between language output and representations, we intuitively hypothesize that such an extra looping process functions as an additional representation-based module, allowing LTs to perceive and monitor the semantics carried by the internal representations. By repeatedly processing their own representation flow, LTs may potentially align their explicit verification process with their superior latent awareness. This report serves as a preliminary verification of this vision, focusing on two inquiries:
\begin{itemize}[leftmargin=1em]
    \item \textbf{Can LTs align linguistic verification output with representations?} We compare the accuracy discrepancies between textual self-verification and representation-based probes across different loop iterations. \textit{While we observe that the gap generally narrows as loops increase, our experiments reveal a nuanced cause: the narrowing gap is partially driven by a performance degradation of the representation probes, rather than solely by an improvement in verbal output.}
    \item \textbf{Can LTs enhance their introspective awareness on representations?} By injecting foreign concepts into the representation during the loop process, we investigate whether the model can effectively recognize and integrate this inserted information. \textit{Contrary to expectations, LTs remains largely insensitive to injections during the intermediate loops, and it is only in the final loop iteration that the injected concept are effectively recognized by the model.}
\end{itemize}
These findings challenge our idealized view of LTs' introspection capabilities. \textbf{The observation in the first inquiry suggests that the loop process, while reorganizing the representations and refining the final verification, may inadvertently reduce the sharpness of LLMs' internal intuition and lead to a loss of representational fidelity. Furthermore, the second inquiry reveals that the model's processing of internal semantics remains local}. Despite the loop architecture, the model does not continuously attend to its internal representations throughout the loop but primarily integrates representational semantic information at the final output stage.

Finally, it is crucial to acknowledge the scope of this study. Our experiments represent a preliminary exploration conducted on a single specific implementation of LTs. Consequently, the observed limitations—such as representational degradation or lack of continuous monitoring on representations—should not be interpreted as intrinsic flaws of the general LTs paradigm. On the contrary, we posit that LTs remain a promising architecture. We believe that with advancements in training objectives and architectural refinements, future iterations can overcome these initial hurdles. We hope that our empirical observations provided in this report will offer valuable insights and inspiration for future research in this domain.

\begin{figure*}[t]
\centering
% \fbox{%
%   \parbox[c][5.5cm][c]{0.95\textwidth}{%
%     \centering \Large Main Figure Placeholder
%   }%
% }
\vspace{-20pt}
\includegraphics[width=1.0\textwidth]{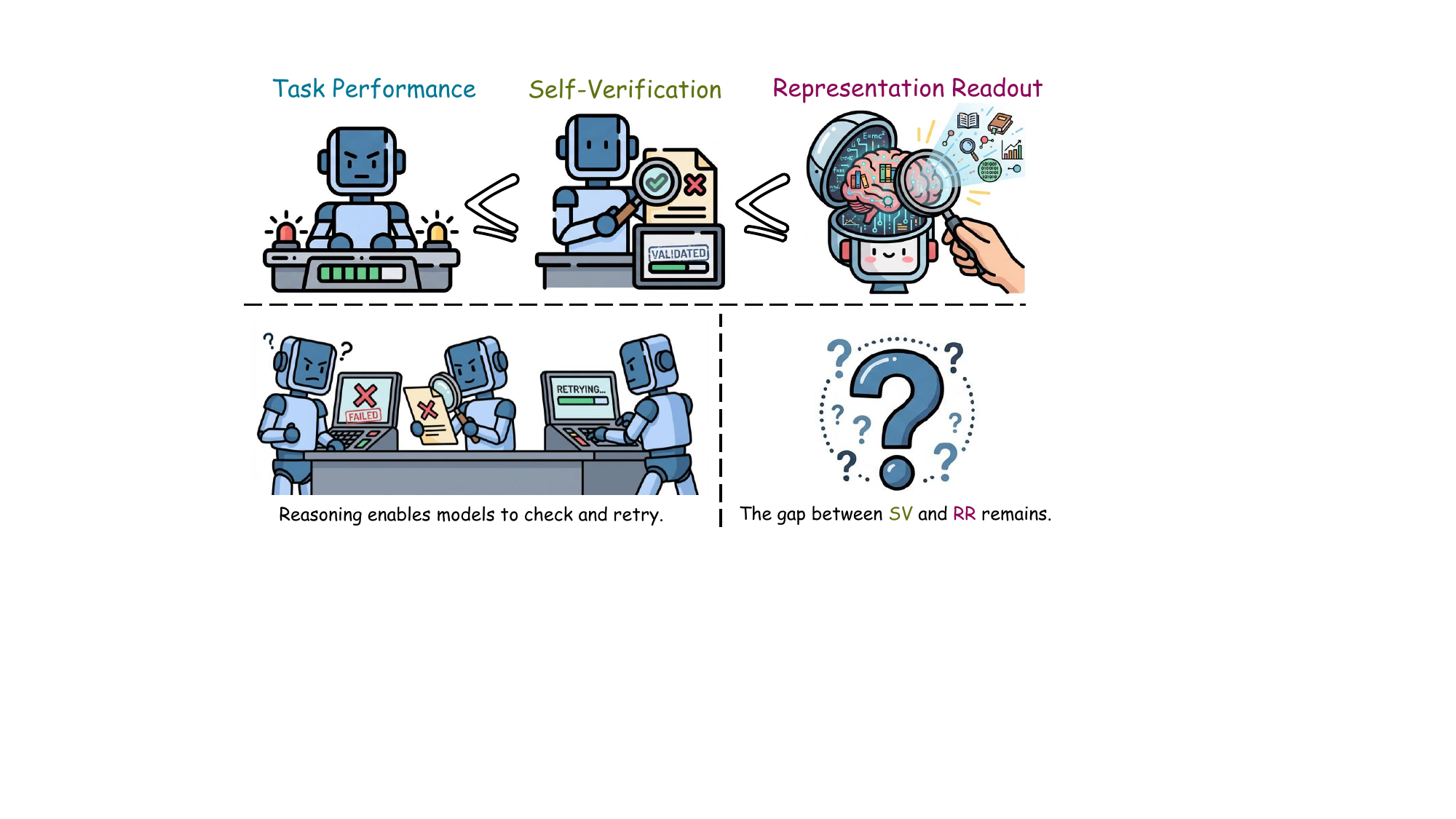} 
\caption{A gap often exists among the three levels of capabilities in practice.}
\label{fig:intro_main_placeholder}
\vspace{-20pt}
\end{figure*}

\section{Preliminary}
\subsection{Looped Transformers}
If Chain-of-Thought (CoT) can be viewed as scaling \emph{sequence length} for more explicit thinking steps, and traditional scaling laws are considered as scaling \emph{parameters and data scope} for stronger capabilities, then Looped Transformers (LTs) open up another scaling dimension: \textbf{scaling of computation depth} \citep{mcleish2025teaching,zeng2025ponderlm,fu2025think,zhu2025scaling}. LTs increase the amount of computation by repeatedly applying the \emph{same} transformer layers to iterate their internal representations and refine their final outputs. Current research on LTs primarily focuses on two parts: How to Loop and When/Where to Loop.

\textbf{How to Loop.} 
Early works explored mechanisms to integrate recurrence into Transformer blocks. \textbf{PonderLM} transforms discrete language prediction into a probabilistic weighted sum of embeddings across iterations \citep{zeng2025ponderlm,zeng2025pretraining}. \textbf{Retrofitting-Recurrence} adopts a modification approach, converting only intermediate layers of standard LLMs into recurrent blocks while keeping shallow and deeper layers as the encoder and decoder, respectively \citep{mcleish2025teaching}. \textbf{THINK-AT-HARD} incorporates LoRA adapters and Duo-causal attention to specifically train LLMs to utilize the additional compute provided by the loop cycles \citep{fu2025think}.

\textbf{Where to Loop.}
A critical challenge in LTs is determining whether to loop or how many loop steps for different tokens. \textbf{MoR} introduces a routing mechanism that dynamically learns how many loop iterations that each token should receive \citep{bae2025mixture}. \textbf{THINK-AT-HARD} studies loop allocation using an oracle policy based on whether an SFT variant of the base model predicts correctly \citep{fu2025think}. \textbf{Ouro} incorporates an early-exit mechanism and employs entropy-regularized objectives during pretraining \citep{zhu2025scaling}. Another specially designed adaptive gating training is employed to teach Ouro when to stop iterating and produce an output.

In this work, we adopt the \textbf{Ouro} series as our primary research object for its superior performance and its compatibility with inference frameworks like vLLM \citep{kwon2023efficient}.

\subsection{AI Monitors}
Current approaches to AI safety typically involve deploying monitors to detect risks, primarily by analyzing LLMs' outputs or inspecting internal representations. LLMs are widely served as an auxiliary AI system (the monitor) to screen the inputs and outputs of another untrusted AI system \citep{madsen2024self,turpin2024language,chen2025reasoning}. In parallel, representation-based monitors like Linear Probes \citep{li2024inference,he2022masked} offer greater reliability, as internal representations are harder for LLMs to fake. In prior work, \emph{in-domain}-trained probes can achieve accuracy that is substantially higher than output-based monitoring \citep{chen2025beyond}. However, this advantage often fails to transfer when evaluated on even slightly shifted data distributions \citep{anthropic2025recommendations}.

\subsection{Introspective Awareness of LLMs}
Anthropic has evaluated introspection by directly manipulating a model's internal activations and then measuring whether the model's self-reports about its own behavior change in ways that correctly reflect those injections \citep{lindsey2025emergent}. They injected activation patterns associated with specific concepts directly into a model's activations and required LLMs to report on their internal representations in various ways. In this report, we conduct experiments on Looped transformers following Anthropic's settings.

\subsection{Formulation}
In this subsection, we consider a task $T$ and an oracle verifier $V$ that determines whether a proposed answer solves the task. Let the policy model $\pi$ produce a solution $A \sim \pi(\cdot \mid T)$, and $V(T,A)=1$ indicates that $A$ successfully completes $T$. 

\subsubsection{Formulation of Hierarchical Performance.}
We study three correct probabilities of different levels, corresponding to (i) Task performance, (ii) Self-Verification, and (iii) Representation Readout.

\textbf{(i) Task performance.}
LLMs' task performance is the probability that its produced answer passes the verifier:
\begin{equation}
\textcolor{ETHPetrol}{P_{TP}}(\pi) \triangleq \Pr_{A\sim\pi(\cdot\mid T)}\!\left[V(T,A)=1\right].
\end{equation}

\textbf{(ii) Self-Verification.}
After producing $A$, LLMs can be prompted to judge whether their own answer is correct with a self-verification strategy $s$. Let $\textcolor{ETHGreen}{SV}_s(T,A)\in\{0,1\}$ denote their self-verification result, where $\textcolor{ETHGreen}{SV}(T,A)=1$ means "I believe my answer is correct."
We quantify self-verification accuracy by agreement with the verifier:
\begin{equation}
\textcolor{ETHGreen}{P_{SV}}(\pi,A,s) \triangleq \Pr\!\left[\textcolor{ETHGreen}{SV}_s(T,A)=V(T,A)\right].
\end{equation}
% Empirically, $\textcolor{ETHGreen}{P_{SV}}$ can exceed $P_A$: even when the model answers incorrectly, it may still be able to detect the error upon reflection.

\textbf{(iii) Representation readout.}
Beyond what LLMs say, we can detect whether their internal representations contain a signal about correctness.
Let $h_l(A)\in\mathbb{R}^d$ denote the model's internal representations (e.g., the residual stream at a chosen layer $L$) when processing $(T,A)$, and a probe $g:\mathbb{R}^d\to\{0,1\}$ be trained to predict the correctness from representations:
\begin{equation}
\textcolor{ETHPurpleDark}{RR}_l(T,A) \triangleq g\big(h_l(T,A)\big).
\end{equation}
We define representation-level correctness readout accuracy as
\begin{equation}
\textcolor{ETHPurpleDark}{P_{RR}}(\pi,A,l,g) \triangleq \Pr\big[ \textcolor{ETHPurpleDark}{RR}_l(T,A)=V(T,A) \big].
\end{equation}

\subsubsection{Formulation of Performance Gaps.}
With these three quantities, we formalize the two gaps discussed in the introduction as follows.
\begin{equation}
\sup_{\pi} \; \textcolor{ETHPetrol}{P_{TP}}(\pi)\;\;\le\;\;\sup_{\pi,A,s} \; \textcolor{ETHGreen}{P_{SV}}(\pi,A,s)\;\;\le\;\;\sup_{\pi,A,l,g}\; \textcolor{ETHPurpleDark}{P_{RR}}(\pi,A,l,g).
\label{eq:upperbound-chain}
\end{equation}
Below we formalize two claim.

\textbf{Claim 1: $\sup_{\pi}  \textcolor{ETHPetrol}{P_{TP}}(\pi)\le\sup_{\pi,A,s}  \textcolor{ETHGreen}{P_{SV}}(\pi,A,s)$.}

The key point is that we can utilize the response to the task as a special case for self-verification.
Formally, for any policy $\pi$, define a degenerate self-verification strategy
\begin{equation}
\textcolor{ETHGreen}{SV}_{s_{\mathrm{deg}}}(T,A) \triangleq 1\{A \text{ is consistent with output}\}.
\end{equation}
That is, self-verification can be carried out by independently re-solving and checking the consistency of results, in which case $\textcolor{ETHPetrol}{P_{TP}}(\pi)\le \textcolor{ETHGreen}{P_{SV}}(\pi,A,s_{\mathrm{deg}})$.
Therefore, optimizing over the larger strategy set cannot yield a worse optimum, giving
\begin{equation}
\sup_{\pi}  \textcolor{ETHPetrol}{P_{TP}}(\pi)\le\sup_{\pi,A,s}  \textcolor{ETHGreen}{P_{SV}}(\pi,A,s).
\end{equation}

\paragraph{Claim 2: $\sup_{\pi,A,s}  \textcolor{ETHGreen}{P_{SV}}(\pi,A,s)\le\sup_{\pi,A,l,g} \textcolor{ETHPurpleDark}{P_{RR}}(\pi,A,l,g)$.}
Self-verification is constrained to express its judgment through LLMs' native language head
(i.e., the particular mapping from internal representations to tokens ).
Thus, if we view $\textcolor{ETHGreen}{SV}$ as a (possibly stochastic) function of some internal representations,
then there exists a (prompt-dependent) mapping $\phi_s$ and a final-layer representation $h_L$ such that
\begin{equation}
\textcolor{ETHGreen}{SV}_s(T,A) = \phi_s\!\big(h_L(T,A)\big).
\end{equation}
Since $\textcolor{ETHPurpleDark}{RR}$ optimizes over all probes $g$ (and even all layers),
it can emulate this choice by selecting $l=L$ and $g=\phi_s$, so
\begin{equation}
\sup_{\pi,A,s} \textcolor{ETHGreen}{P_{SV}}(\pi,A,s)
= \sup_{\pi,A,s}\Pr\!\left[\phi_s(h_L)=V\right]
\le \sup_{\pi,A,l,g}\Pr\!\left[g(h_l)=V\right]
= \sup_{\pi,A,l,g} \textcolor{ETHPurpleDark}{P_{RR}}(\pi,A,l,g).
\end{equation}

\textbf{Summary.} Consequently, it is natural that the \emph{best possible} achievable accuracies satisfy above claims, even though for a particular fixed LLM and fixed prompts one may not always observe strict inequalities.

\section{Can LTs align linguistic output with representation?}

\subsection{Setups.}
We evaluate LLMs' language-based self-verification ability and representation-based monitor classification accuracy in two scenarios: (1) \textbf{safety judgment} and (2) \textbf{mathematical verification}. We examine how the gap between them changes with the number of loop steps. For language-based self-verification, we only use prompts to ask LLMs to make judgments and evaluate it on the test set. For the representation-based monitor, we first extract representations from the layer at 80\% depth on the training set to train linear probes, and then evaluate them on the representations from test-set.

In the safety scenario, we use the BeaverTails dataset \citep{ji2024beavertails}, collecting 8,000 safe and 8,000 unsafe QA pairs as the training set, and 1,000 safe and 1,000 unsafe QA pairs as the test set. In the math scenario, we use the DeepMath dataset \citep{he2025deepmath}: for each problem, we rollout ten responses for each question with Qwen3-235B-A22B-Instruct-2507 \citep{yang2025qwen3}. We then sample 8,000 correct and 8,000 incorrect answers as the training set from problems with an accuracy of 40\%–60\%, and 2,000 correct and 2,000 incorrect answers as the test set.

We conduct experiments on the 1.4B and 2.6B versions of the Ouro model and their reasoning variants with loop steps from 1 to 8 \citep{zhu2025scaling}. On standard Transformer LLMs, we run the same experiments as baselines including Llama3.1-8B-Instruct \citep{dubey2024llama}, Qwen3-4B-nothink \citep{yang2025qwen3}, Qwen3-8B-nothink, and Qwen3-30B-A3B-Instruct-2507. We report the accuracy and F1 score as evaluation metrics, and quantify the gap as the performance of the representation-based monitor minus that of language-based self-verification (\textit{i.e.}, Metric(\textcolor{ETHPurpleDark}{RR}) - Metric(\textcolor{ETHGreen}{SV})).

\begin{figure}[!t] % [htbp] 是位置参数，意为这里(h)、顶部(t)、底部(b)或单独一页(p)
    \vspace{-10pt}
    \centering % 让图片居中
    % width 可以设置宽度，例如 0.8\textwidth 表示占页面宽度的80%
    % {filename} 是你的图片文件名（如 image.png 或 figure1.jpg）
    \includegraphics[width=1.0\textwidth]{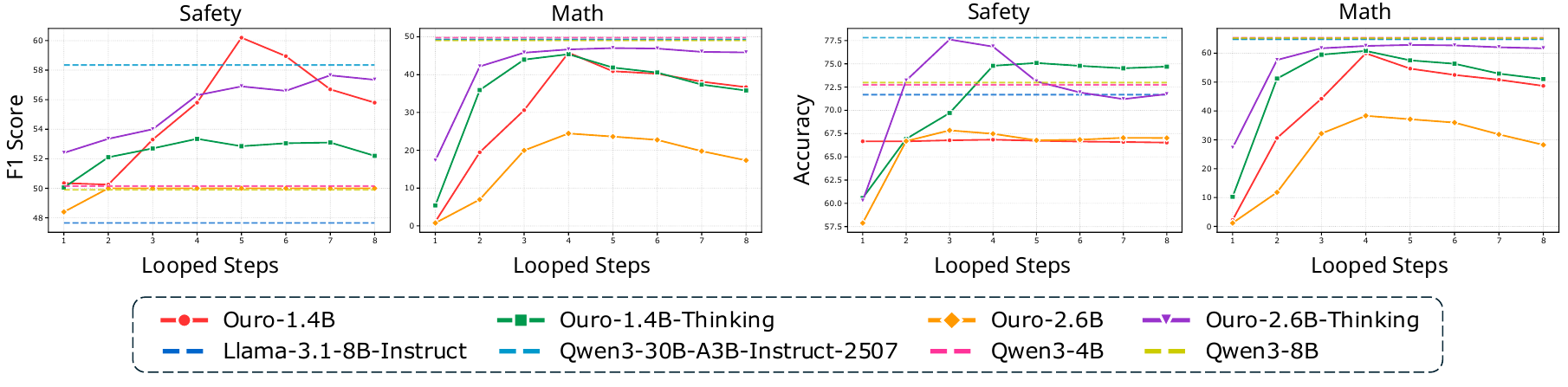} 
    \vspace{-15pt}
    \caption{Accuracy/F1 of language-based self-verification.} % 图片下方的说明文字
    \label{fig:cot} % 【关键】这是图片的唯一标签，引用时就用这个名字
\end{figure}

\begin{figure}[!t] % [htbp] 是位置参数，意为这里(h)、顶部(t)、底部(b)或单独一页(p)
    \vspace{-10pt}
    \centering % 让图片居中
    % width 可以设置宽度，例如 0.8\textwidth 表示占页面宽度的80%
    % {filename} 是你的图片文件名（如 image.png 或 figure1.jpg）
    \includegraphics[width=1.0\textwidth]{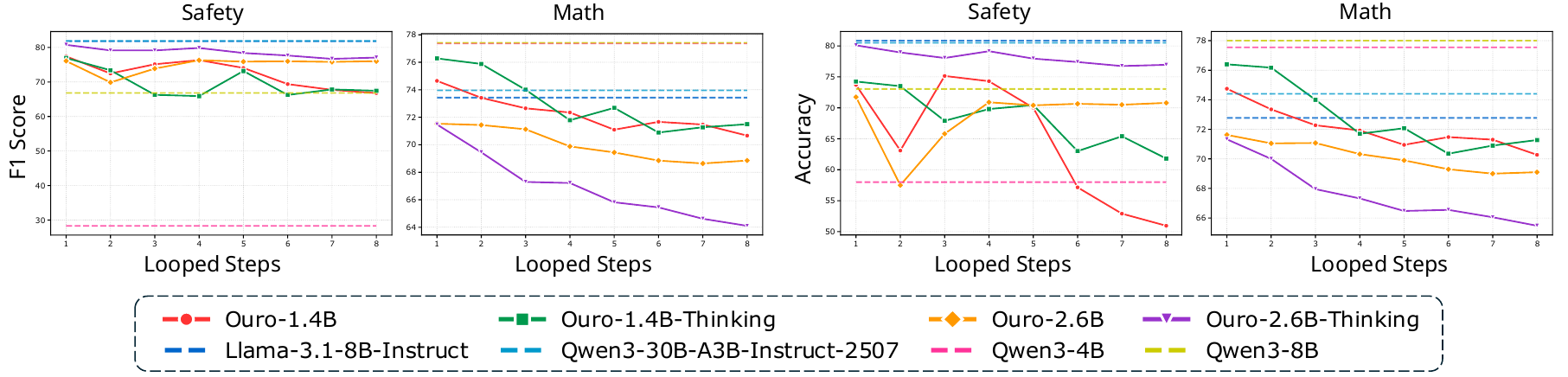} 
    
    \vspace{-5pt}
    \caption{Accuracy/F1 of linear probes trained on internal representations.} % 图片下方的说明文字
    \label{fig:lp} % 【关键】这是图片的唯一标签，引用时就用这个名字
\end{figure}

\begin{figure}[!t] % [htbp] 是位置参数，意为这里(h)、顶部(t)、底部(b)或单独一页(p)
    \vspace{-10pt}
    \centering % 让图片居中
    % width 可以设置宽度，例如 0.8\textwidth 表示占页面宽度的80%
    % {filename} 是你的图片文件名（如 image.png 或 figure1.jpg）
    \includegraphics[width=1.0\textwidth]{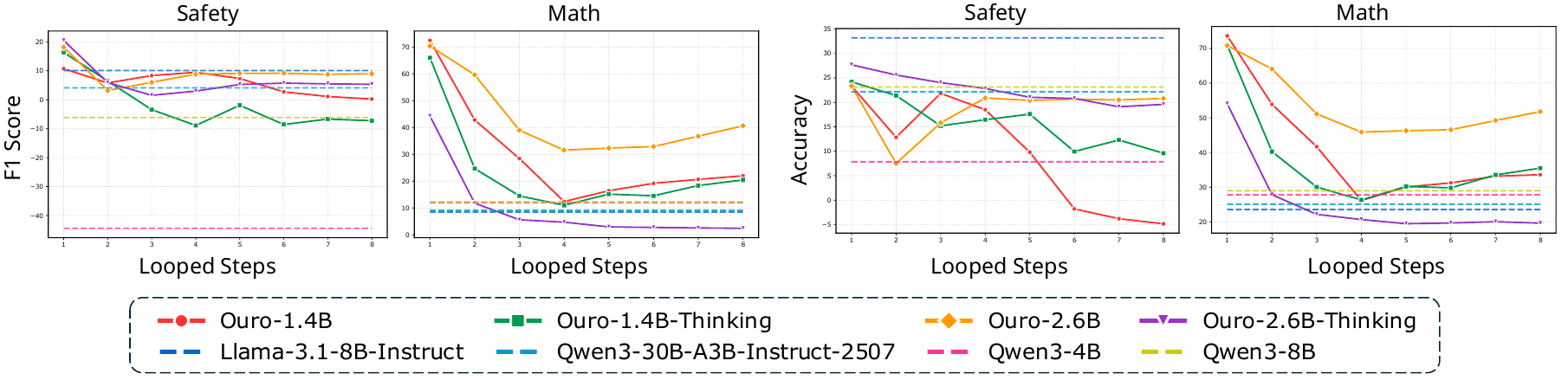} 
    
    \vspace{-5pt}
    \caption{Gap between representation readout and language-based self-verification.} % 图片下方的说明文字
    \label{fig:gap} % 【关键】这是图片的唯一标签，引用时就用这个名字
    % \vspace{-20pt}
\end{figure}

\subsection{Results.}
As shown in Fig.~A, in both scenarios the accuracy of textual verification exhibits an overall upward trend as the number of loops increases. This is consistent with the intuition that scaling computational depth leads to stronger capabilities. However, Fig.~B shows that the performance of the representation-based linear probe declines slowly across increasing loop iterations. This suggests that the information contained in the representations does not become more separable with additional loops; instead, it may be partially degraded, making it harder for the monitor to detect.

From Fig.~C, we observe that the performance gap indeed decreases as the number of loops increases, aligning with our expectation stated in the introduction. However, this narrowing gap is not solely driven by improved textual verification performance but also partly attributable to the decline of representation-based monitoring performance. If the cost of looping is reduced representational readability (\textit{i.e.}, lowering the ceiling $\textcolor{ETHPurpleDark}{P_{RR}}$ and thereby ``aligning downward''), then this may not be the kind of improvement we ultimately desire.
\section{Can LTs enhance their introspective awareness on representations?}
\subsection{Setups.}
We follow the setups used in \cite{lindsey2025emergent} to evaluate the LTs' introspective awareness on representations. Specifically, we extract the \emph{concept vector} for 50 concepts using the difference vector between their representations and an background representation (calculated by the average of another 100 concepts) from the layer at 80\% depth. Then, we ask LTs if they can detect an injected thought and what is the injected thought about, as the same time as we inject the concept vectors into their representations. 

We use Qwen3-235B-A22B-Instruct-2507 to evaluate the responses of LTs, assessing (i) whether the model is aware of the injection and (ii) whether it can accurately identify the injected concept. Based on these judgments, we consider two metrics: (1) the adjusted detection accuracy--the detection accuracy with real injection minus the hallucination detection accuracy when no injection is applied; and (2) the accuracy of correctly answering the injected concept. For each concept, we run 64 trials and report the average results. 

To quantify the contribution of the scaling of loops, we extract concept vectors from LTs with 1--8 loops, and, during generation with 1--8 loops, we conduct injection experiments at every loop iteration of them. In this way, for each model and metric, we visualize the results as 8 $8\times 8$ matrix, where each entry corresponds to injecting a concept vector extracted at an arbitrary loop step into an arbitrary loop step during generation.

\begin{figure}[!t] % [htbp] 是位置参数，意为这里(h)、顶部(t)、底部(b)或单独一页(p)
    \vspace{-20pt}
    \centering % 让图片居中
    % width 可以设置宽度，例如 0.8\textwidth 表示占页面宽度的80%
    % {filename} 是你的图片文件名（如 image.png 或 figure1.jpg）
    \includegraphics[width=1.0\textwidth]{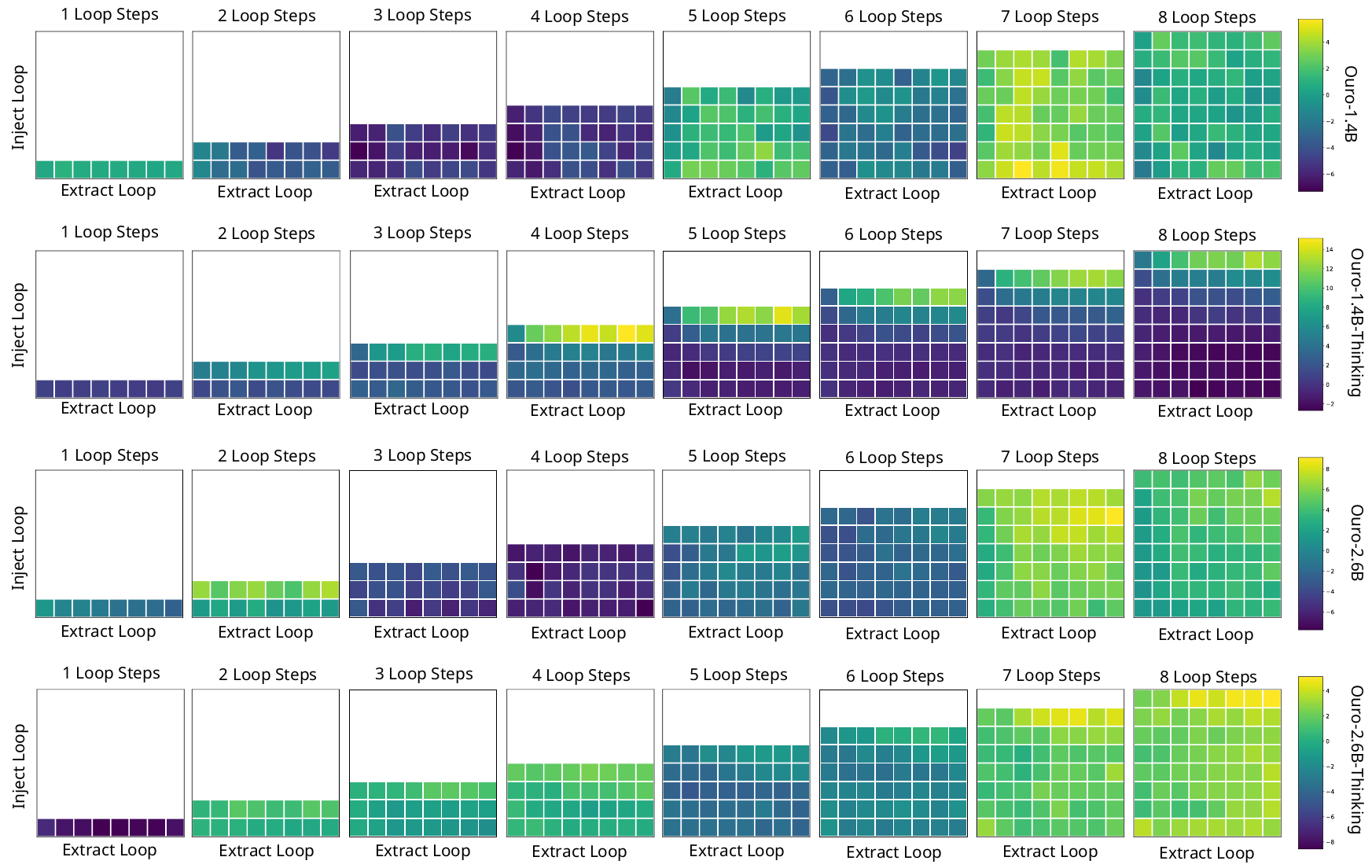} 
    
    \vspace{-5pt}
    \caption{Injection awareness across the extract loop and inject loop.} % 图片下方的说明文字
    \label{fig:detect} % 【关键】这是图片的唯一标签，引用时就用这个名字
\end{figure}

\begin{figure}[!t] % [htbp] 是位置参数，意为这里(h)、顶部(t)、底部(b)或单独一页(p)
    \vspace{-20pt}
    \centering % 让图片居中
    % width 可以设置宽度，例如 0.8\textwidth 表示占页面宽度的80%
    % {filename} 是你的图片文件名（如 image.png 或 figure1.jpg）
    \includegraphics[width=1.0\textwidth]{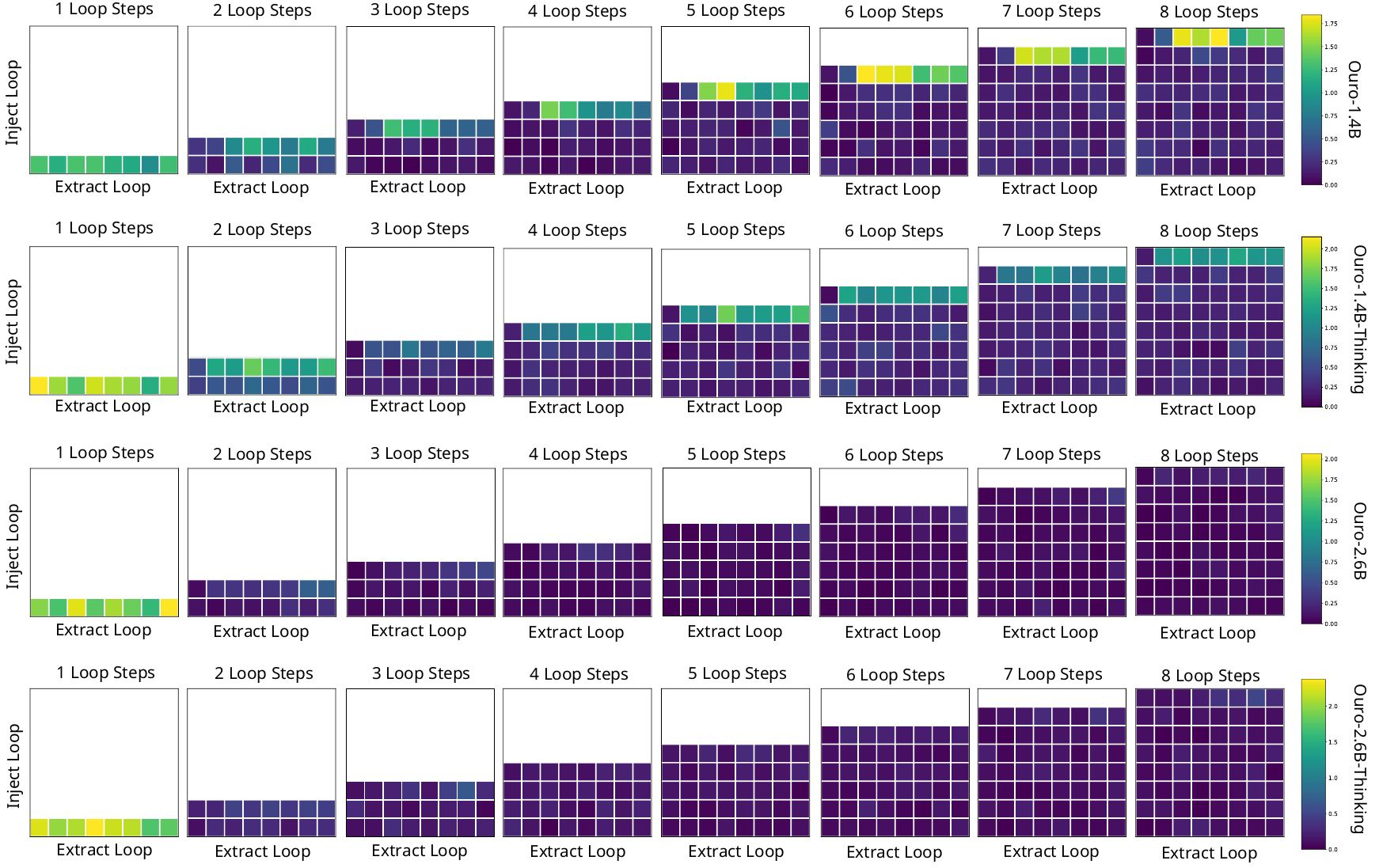} 
    
    \vspace{-5pt}
    \caption{Injected concept identification across the extract loop and inject loop.} % 图片下方的说明文字
    \label{fig:identify} % 【关键】这是图片的唯一标签，引用时就用这个名字
    \vspace{-20pt}
\end{figure}
\subsection{Results.}
As shown in Fig. D, the adjusted detection accuracy exhibits an upward trend as the number of loops increases. However, concept vectors injected closer to the final loop are more likely to be noticed. Similarly, Fig. E shows that the accuracy of correctly identifying the injected concept improves only when the concept is injected in the final loop. This indicates that representations injected in earlier loops do not become easier to recognize in the language output as the preceding loops accumulate. This is inconsistent with our expectation of "continuous introspection”: even multiple loops are applied, the semantic processing scope of LTs over the internal representations still seems to be limited to the last loop.

\section{Conclusion}
This report investigates whether Looped Transformers can bridge the gap between internal representations and linguistic outputs by utilizing their recursive structure as an iterative introspective mechanism. Our experiments reveal that while increasing loop iterations narrows the accuracy gap between self-verification and representation probes, it is partly driven by a degradation in the performance of representation-based probes. Furthermore, we observe that the injections of concepts are detected and identified mainly when injected in the final loop, suggesting that current LTs, at least for the Ouro instantiation studied here, does not yet yield introspective awareness through early and intermediate loops.

\bibliography{iclr2026_conference}
\bibliographystyle{iclr2026_conference}

% \newpage
% \appendix
% \section{Appendix}

\end{document}

%% file: math_commands.tex
\usepackage{amsmath,amsfonts,bm}

\def\eqref#1{equation~\ref{#1}}
\def\1{\bm{1}}

\DeclareMathAlphabet{\mathsfit}{\encodingdefault}{\sfdefault}{m}{sl}
\SetMathAlphabet{\mathsfit}{bold}{\encodingdefault}{\sfdefault}{bx}{n}